Chapter 10

**Semantic Vector Spaces for Broadening Consideration of Consequences**

**Douglas Summers Stay**

Reasoning systems with too simple a model of the world and human intent are unable to consider potential negative side effects of their actions and modify their plans to avoid them (e.g., avoiding potential errors). However, hand-encoding the enormous and subtle body of facts that constitutes common sense into a knowledge base has proved too difficult despite decades of work. Distributed semantic vector spaces learned from large text corpora, on the other hand, can learn representations that capture shades of meaning of common-sense concepts and perform analogical and associational reasoning in ways that knowledge bases are too rigid to perform, by encoding concepts and the relations between them as geometric structures. These have, however, the disadvantage of being unreliable, poorly understood, and biased in their view of the world by the source material. This chapter will discuss how these approaches may be brought together in a way that combines the best properties of each for understanding the world and human intentions in a richer way.

## 10.1 Designing for Safety

Failure Mode and Effects Analysis documents are used for ensuring safety in complex systems such as automotive design. Engineers painstakingly analyze each subsystem for its probability of failure and build in layers of redundancy depending on the seriousness of system failure. Fail-safes (systems that, when they fail, do so in a way that leaves them safer), layers of redundancy, and hazard and risk analysis, are all tools used to reduce the probability of injury or death to a reasonably low level.

Typical machinery makes use of a very simple model of the world. A grocery store conveyor belt, for example, has two states and one binary sensor controlling which state it is in. A safety analysis would consider a richer model of the conveyor belt as a collection of moving parts, any one of which could break, and the much larger set of states that could put the system in, as well as potential consequences of such a failure. The complexity of autonomous systems makes such analysis more difficult. As the system becomes more autonomous, the number of potential actions the system can take and the variety of situations it can find itself in grows very quickly. In addition, useful AI systems must learn and change over time: understanding means incorporating newly acquired facts about the world into the already existing body of knowledge. A-priori consideration of every possible situation becomes impossible. It seems the only solution is to automate the safety analysis itself: we must design the system to perform a safety analysis on its own actions.

Doing this would require the autonomous system to have a rich model of the entire environment it will be interacting with—not just a simplified model that allows it to perform its normal tasks, but a model that takes into account the wider environment so that it understands what its tasks are for, what consequences its actions will have, and which consequences are to be avoided.

Creating such a system to reduce human error would be very difficult, difficult to the point that it has never been seriously attempted. Causal reasoning about physical systems can be performed for limited situations by creating detailed physical simulations, such as finite element analysis for stress analysis or nuclear weapons testing, but such methods are far too computationally intensive to be used for making quick decisions about everyday situations. A more promising approach involves qualitative reasoning about physical systems. In 1985, the dramatically named "Naïve Physics Manifesto" (Hayes 1985) laid out a program for enabling AI to answer questions about real world situations, with some initial success: "figuring out that a boiler can blow up, that an oscillator with friction will eventually stop, and how to say that you can pull with a string, but not push with it." Hayes' plan involved entering knowledge about the causal relationships of physical systems into a first order logical system (a knowledge base), and deducing answers to such questions. This approach ran into the common problem of expert systems: brittleness and incompleteness (Lenat, 1985). Unless a query was designed carefully by a researcher with intimate understanding of how the knowledge base was constructed and what information it contained, some missing assumption would break the chain of reasoning and no answer would be returned.

There has been substantial work (e.g., Dash, 2013) on A.I. planning and the creation of subgoals. While this is important and necessary, as long as these subgoals make use of a simple, incomplete model of the world, they will be inherently unsafe outside of toy applications.

**10.2 Understanding Intent**

Reasoning about physical processes that may lead to accidents, while a huge effort in itself, is only one part of the problem. Without understanding exactly the goal to be accomplished, the AI system may plan for a goal in a way that contradicts other implicit goals, in ways that may prove dangerous.

Amodei (2016) pointed out two mechanisms that can lead two accidents when an objective function is specified. "Negative side effects" can occur because of an insufficient model of chains of causal relations, leading to unanticipated negative consequences. "Reward hacking," however, occurs when the objective function is technically satisfied, but in a way that contradicts unspoken goals.

"The Sorcerer's Apprentice" is an old story, probably most familiar from Disney's version, but originating in the second century A.D. The ancient Greeks also told stories of King Midas turning his daughter to gold or Tithonus, who Zeus grants immortality but not eternal youth. There are similar stories about genies from the Arabian nights, as well as fairy tales about wish-granting fishes, stories of golems from Jewish sources, and stories of deals with Old Nick from frontier America. All these fit the same pattern. In each version of the story, the entity granting the petition has the ability to help humans achieve their goals, but although the petitioner's goal is technically satisfied, it happens in a way that contradicts real, deeper desires. In discussing issues of A.I. safety, Stuart Russell points out that we have a similar situation: "The primary concern is… simply the ability to make high-quality decisions. Here, quality refers to the expected outcome utility of actions taken, where the utility function is, presumably, specified by the human designer….The utility function may not be perfectly aligned with the values of the human race, which are (at best) very difficult to pin down." (Russell 2014)

Dietterich (2015) wrote, "Suppose we tell a self-driving car to 'get us to the airport as quickly as possible!' Would the autonomous driving system put the pedal to the metal and drive at 125 mph, putting pedestrians and other drivers at risk? … [T]hese examples refer to cases where humans have failed to correctly instruct the AI system on how it should behave. This is not a new problem. An important aspect of any AI system that interacts with people is that it must reason about what people intend rather than carrying out commands literally. An AI system must analyze and understand whether the behavior that a human is requesting is likely to be judged as "normal" or "reasonable" by most people."

This is a familiar experience to every programmer. Although programming languages allow us to specify exactly what we want the computer to do, we often end up writing buggy programs that don't do what we actually want. Autonomous systems are designed to act with less direct, more natural instruction. How do we make a system that will carry out what we want when we ask it to do something? It is impossible unless the system has knowledge of what kinds of things we want and what our words mean.

The problem of A.I. safety, then, is inescapably a version of the same problem of automating understanding that lies at the core of natural language understanding, common sense reasoning, mental modeling, creativity, and many other efforts that have been challenges for A.I research since its inception. This can be looked at in a positive, way, though. The same research that is required to make A.I. effective at real world tasks will also be advancing the ability to carry out those tasks safely, without undesirable side effects.

## 10.3 Expressing intent

Natural languages, unlike programming languages, are imprecise and underspecified. In every uttered sentence, there is a large body of assumed background context, shared knowledge that can remain unsaid. Part of this is innate: all humans have certain shared goals even from infancy, such as air, water, food, and safety from physical and emotional harm—Maslow's "hierarchy of needs." Part of this is learned over a lifetime, the cultural body of knowledge such as property rights, social conventions, sarcasm, humor, and so forth.

When a command is expressed in natural language, the command cannot contain all of the limitations and context necessary to carry out the command in a way that matches the intent of the human giving the command. Such precision in language is inherently *un*-natural. If not expressed in the command itself, such values must already be included in the background knowledge brought to bear as the A.I. forms a plan to carry out the command.

One well-established attempt at pinning down some part of human values is the legal system. The legal system attempts to encode some human standard of what is acceptable behavior of an agent interacting with society and the world in very precise language, at least as far as human-readable documents go. The written law, however, is insufficient to decide cases. When cases are actually brought to trial, human lawyers are needed. The lawyers' role is to search through similar cases which have already been decided, in order to find the nearest analogies with cases in precedent they can which result in a ruling favorable to their side. With lawyers performing this role on each side of the case, a human judge or jury decides which they find to be most similar.

Human judges and lawyers are needed because the law is necessarily insufficiently precise to cover every possible case. In this way it is very similar to hand-created knowledge bases. Attempts to encode knowledge in a system capable only of deductive reasoning were invariably very limited in their usefulness, because they lacked this ability to extend reasoning to new cases by analogy (Speer 2008).

The ability to find analogies is essential to understanding physical systems and human intent. Suppose a boy hits a baseball through a window of the house. A mother provides negative reinforcement, saying "don't do that again." But what does she mean by "do that"? It could mean:

- "never hit a ball with a bat"
- "don't play near the house"
- "don't hit a ball towards this particular window"
- "don't move your arms"

And so forth. In order to understand what she means by "do that," the boy may apply the golden rule: his unconscious reasoning is something like, "I would be angry if someone broke one of my possessions, so she must be angry because I broke one of her possessions." The boy will recognize that throwing a stick inside the house where it might break a lamp is an analogous situation to be avoided in the relevant sense of "causing an object to move unpredictably where it has the chance to damage someone's fragile property." But the ability to pull out this particular meaning of "do that" over any of the others depends on a lifetime of experience and an internal set of desires that corresponds, more or less, with the mother's.

Is building this kind of "common sense" into an AI system really necessary for it to behave safely? It is such a difficult problem that any way around it seems preferable. In Amodei (2016), several methods were proposed for increasing AI safety that don't explicitly include such a design. For example, they suggested avoiding side effects, or situations which might potentially have side effects. After exploring this idea for a little while, however, they pointed out situations where such an approach would fail without some notion of the user's goals and the form that consequences would take. There's no free lunch: (Amodei 2016, p. 6) "Avoiding side effects can be seen as a proxy for the thing we really care about: avoiding negative externalities. If everyone likes a side effect, there's no need to avoid it. What we'd really like to do is understand all the other agents (including humans) and make sure our actions don't harm their interests…However we are still a long way away from practical systems that can build a rich enough model to avoid undesired side effects in a general sense." The only solution to this problem is the hard one: biting the bullet and building a rich enough model to avoid undesired side effects.

There are two main problems with encoding such common sense background knowledge in a way that an autonomous system can make use of. The first problem is an architectural issue: The meanings of concepts are rich and nuanced. What kind of data structure can allow for such diverse phenomena as being reminded by similar ideas, completing analogies, recognizing objects by their attributes, and recognizing a class by a single example of that class, and still support deductive reasoning?

The second problem is this: once we have an architecture capable of storing concepts and reasoning about them in deductive, inductive, and analogical ways, how can we populate it with the vast amount of common-sense knowledge we all share?

**10.4 Problem 1: An Encoding For Concepts**

Douglas Hofstadter has been writing about the nature of concepts and analogies since the

1980s, pointing out a distinction between how symbolic information is stored in precise logical forms in a knowledge base, and how concepts are held in the mind. "The property of being a concept is a property of connectivity, a quality that comes from being embedded in a certain kind of complicated network." (Hofstadter 1985, p. 528) In an object-oriented programming language or a knowledge base, we can represent an object such as a fire-extinguisher with a few facts defining its function as needed in the program. To really count as a *concept*, though, requires much more than that. The concept of a fire-extinguisher includes something of its shape and size, the material it's made from, its appearance, the uses it is put to, how to operate it, where one can be found, a rough idea of how much it costs, what it resembles, and many other such properties. Each of those properties, in turn, must be concepts, with the same richness of internal structure. Concepts that define a class have shades of membership. A bucket of sand might be considered a fire-extinguisher under certain ill-specified conditions. A fire-extinguisher that has not been recharged also has a shaded inclusion in the category.

Concepts have connections of varying strength with many other concepts. "Each new concept depends on a number of previously existing concepts. But each of those concepts depended, in its turn, on previous and more primitive concepts… This buildup of concepts over time does not in any way establish a strict and rigid hierarchy. The dependencies are blurry and shaded rather than precise, and there is no strict sense of higher and lower… since dependencies can be reciprocal. New concepts transform the concepts that existed prior to them, and that enabled them to come into being; in this way, newer concepts are incorporated inside their "parents" as well as the reverse." (Hostadter 2013, p. 54) To act as a concept, then, requires that the information be stored in a way that admits degrees of similarity, and definitions that are reciprocal, rather than built up from axioms like the definitions of mathematical structures.

Our understanding of concepts is evoked by similar concepts, and the way we think about concepts is largely analogical in nature. "The ability to perceive similarities and analogies, he argues, is one of the most fundamental aspects of human cognition. It is crucial for recognition, classification, and learning, and it plays an important role in scientific discovery and creativity." (Vosniadou 1989, p. 1). Whatever representation of concepts we come up with, it must be able to support reasoning by analogy, and such analogies must be flexible enough to admit ambiguity and imperfect matches.

In early A.I. research, concepts were represented using knowledge bases: as nodes in a relational graph in a database with the capacity for deductive reasoning. The graph expressed first order relations as connections between nodes. This stored symbolic information, but failed to capture the subsymbolic information that is inherent in human concepts. A key problem in storing information this way is that any mismatch between the arrangement of concepts in the knowledge base and the form of a query will cause the query process to fail completely, returning no results at all. For example, the knowledge base may include the fact that gasoline may catch fire:

**causes (gasoline, fire_hazard)**

but a query asking

**has_tendency (gasoline, X?)**

Will return no results unless the knowledge base also has rules defining how **causes** and **fire_hazard** are connected to **has_tendency** and **burn_rapidly**.

This isn't just a problem with insufficient rules in the knowledge base, however. Concepts in the human brain seem to be stored in a way that makes them fundamentally different from entries in a knowledge base. We can be reminded of concepts by resemblance in sounds between words, similar parts, or properties between concepts, a similar environment in which the concepts are encountered, and many other ways. Instead of being a discrete graph where each concept in the graph is assigned or not assigned to a particular relation, there are gradations of inclusion by which a pair of terms fits the relation more or less precisely. Many of the relations we can find in our memory seem to be an implicit result of the way the concept is stored, rather than an explicitly learned link.

(Kanerva 1988, p. 2) wrote "although we normally ignore such links, they are there, and they can tell us something about the mathematical space for memory items. Translated into a requirement for the model, memory items should be arranged in such a way that most items are unrelated to each other but most pairs of items can be linked by just one or two intermediate items. This requirement affects the choice of mathematical space for memory items, also called the semantic space."

## 10.5 Semantic Vector Spaces

What Kanerva suggested was to encode the concepts as vectors in a high-dimensional vector space. High-dimensional vector spaces have some unintuitive properties that make them ideal for representing concepts. One of the most important of these is that between two arbitrary vectors in this space, we can find a vector very close to both but not unusually close to any unrelated vectors. The vector spaces Kanerva worked with were n-dimensional binary vector spaces, where each element of the vector is 1 or 0, written as $\{0,1\}^n$. "The distance between two points of $\{0,1\}^n$ represents the similarity of two memory items—an association based on form. It is the number of places in which the two patterns differ, so that the closer the points, the more similar the items. Almost all of the space is nearly indifferent to (or about $n/2$ bits away from) any given point, whereas two points $n/4$ bits apart are very close together in the sense that an exceedingly small portion of the space lies within $n/4$ bits of a point. This is intuitively appealing in that any particular concept in our heads is unrelated to most other concepts, but any two unrelated concepts can be linked by a third that is closely related to both." (Kanerva 1988, p. 25)

Starting with a few primitives, a high-dimensional vector space can build them up to represent more complex ideas. For example, the vector representing **ice** can be located near the sum of the vector for **cold** and the vector for **water**. This new **ice** vector will be similar to both **cold** and **water** and nothing else, except any other terms we may have also formed that include **cold** and **water** as components, such as **snow**. To count as a concept, the vectors would need to be built from many more components representing every aspect of ice that might be of interest, but with a high-enough dimensional vector, many such components can be included. Details about how many such components can be included in a single vector can be found in (Kanerva 1988) or (Hawkins 2004).

Such a semantic vector space is capable of representing not just ideas, but also relations

between them. For example, suppose we wanted to represent the fact that snow causes icy roads:

**causes_road_condition(snow, icy_roads)**

We define locations in the vector space representing the concepts **precipitation**, **frozen**, and **road**. To represent **snow**, we take the sum of **precipitation** and **frozen** and to represent **icy_roads** we sum **frozen** and **road**. The relation **causes_road_condition** is then the vector which subtracts out **precipitation** and adds in **road** to a concept: the vector **(road - precipitation)**. This same relation vector, when added to **rain** will lead us to the vector for **wet_roads**.

In this way, one-to-one relations between concepts can be defined as displacement vectors between the vectors for those concepts. Concepts can be built up from the simplest attributes we wish to define. In a real system, we would, of course want a more refined concept for **icy_roads** that included the fact that they are slippery, that they sometimes have a reflective appearance, and so forth. The problem of how to get all of the information that needs to be encoded in a concept will be covered in section 10.6. All we are doing here is showing that the vector space representation is capable of holding such information about concepts and their relations.

Following chains of deductive reasoning would be simple in such a vector space. Suppose the space encodes the facts that

**has_location (finger , cutting_board)**

and

**uses (cutting_board, knife)**

then we can conclude that **could_be_affected_by (finger, knife)** using a rule stating that **has_location(X,Y) ^ uses (Y, Z)** implies **could_be_affected_by (X,Z)**. (Neelakantan (2015) explores such chained reasoning in vector spaces.) In this case, the vector representing **could_be_affected_by** can be found by simply adding the vectors from **X** to **Y**, and from **Y** to **Z** to find the vector from **X** to **Z**.

It is also possible to perform analogical reasoning in such a vector space. Suppose we are given the following analogy to solve: **bear:hiker::shark:X**. If the concepts **bear**, **hiker**, and **shark** are already in the vector space, they are composed of simpler terms. Suppose these simpler terms happen to be **woods**, **sea**, **predator** and **tourist**. Then substituting in the simpler component terms, we have the simpler analogy **woods + predator : woods + tourist :: sea + predator : X**. The relation between the first two terms can be found by subtracting **predator** from **bear**, (leaving **woods**) and adding **tourist** to the result. We then apply that relation to **shark**, to get **sea + tourist**, which is close to the vector for **snorkeler**. The vector arithmetic simplifies to **D = B + C – A**, when trying to solve **A:B::C:D**. While these are too high-level terms to actually be used as primitive concepts, they serve to demonstrate the arithmetic.

To the extent that the fundamental concept vectors are plentiful enough, this vector space also serves as an associational memory, in the sense that summing up a few related terms is enough to bring to mind a term associated with them. For example, adding **France** + **city** +

**fashion** gives a vector close to Paris, since the components of **France**, **city**, and **fashion** all added together are similar to the components of **Paris**, plus some leftover that can be treated as noise.

Thus, a memory encoded as a high dimensional vector space is capable of supporting deductive, analogical, and associational reasoning. As further support for the practicality of such an approach, it is interesting to note that such a memory is fairly biologically plausible. As a toy model, each component of the vector could be considered to be a neuron which is activated to some degree, and reasoning would consist of bringing to mind the various concepts in such a way that the end result of the reasoning is the application of the relevant vector arithmetic on those concepts. Hinton (1984) outlined how distributed representations were more biologically plausible than the local representations used in a knowledge base. Brain-imaging studies have likewise suggested that concepts are represented in the brain as distributed networks of neural activation (Rissman 2012), (Blouw 2005). In particular, the analogical-reasoning capability of a semantic vector space can be understood as an example of the relational priming model of analog- making outlined in Leech (2008). There is also evidence that object categories are encoded as a continuous semantic space across the surface of the brain (Huth 2012). The brain's slow operating speed and massive parallelism also hint that whatever operations are being performed must be very short, simple programs operating on large vectors.

All this is of little use, however, unless we can find a way to input all the information about the concepts. Indeed, unless we can find some way of automating the process of populating the vector space, we are not getting much more out from it than we have painstakingly entered by hand. Once we know that two concepts share certain qualities or relations, finding associations and analogies is not difficult. It is recognizing those shared properties in the first place that is the harder problem.

This is where the problem stood for some time after the development of the theory of representing concepts as high-dimensional vectors. At that time, finding a way to automatically populate the vector space was not a practical possibility. The development of such a method would come about from attempts to find semantically similar documents.

**10.6 Problem 2: Distributional Semantic Vector Spaces**

A distributional semantic vector space assigns a vector to each word such that words found in similar contexts have similar vectors. Incredibly, this is enough to create (in an approximate, noisy way) the kind of vector space described in the previous section, including some very subtle and difficult to express attributes of concepts that can be used for associational and analogical reasoning.

The simplest distributional vectors represent the meaning of a document by counting up the frequency of occurrence of each word in that document. The vector is the size of the entire English vocabulary **v**, with zeros for most words, which never occur in the document, and the occurrence count for the words which do occur. Such vectors are impractically large, and tend to be very noisy in terms of similarity between documents expressing similar ideas, because one author may prefer certain terms to express an idea, while another author would use a different subset of terms to express the same idea.

If we consider a few words to either side of a given word any time it occurs to be its "document", we can create a v * v matrix in this way that encodes each word by its context. Using support vector decomposition (SVD), we can reduce this matrix to a more reasonable size (say, 300 x 300 instead of **v** x **v**) and at the same time remove much of the noise, so that vectors encoding similar terms end up with similar vectors. Later techniques such as word2vec optimized certain technical parameters and improved the time and memory performance of deriving such vectors (actually performing SVD on a **v** x **v** matrix requires an impractical amount of memory) to the point where enormous corpora could be practically handled, but the core idea behind the vectors is the same (Deerwester 1998). Since this process assigns vectors with similar meanings similar vectors, all the nice properties listed above—the ability to find analogies, to recall a concept by its associations, to break a concept down by its attributes, to encode a one-to-one relation as a vector, and to compose these vectors to follow chains of reasoning—are *automatically* properties of the distributional vectors.

Consider the analogy

**seat_belt : car :: life_preserver : X**.

**seat_belt** will be found in contexts discussing roads, and in contexts discussing accidents and safety. **Life_preserver** will also be found in contexts discussing accidents and safety, but instead of being found in contexts having to do with roads, it will be found much more often with words like **sea** and **ocean**. **Car** will be found in contexts having to do with vehicles, as well as contexts having to do with roads. **Ship** and **boat** will also be found in vehicle contexts, near verbs like **travel** which it shares with **car**, for example, but also in the context of **ocean**. Since a word found in two concepts is found near the average or sum of those two concepts, the following will be approximately true:

**road + safety : road + vehicle :: ocean + safety : ocean + vehicle**

This is the same kind of analogy we saw in the constructed example of sharks and bears above. This conceptual arithmetic would not be exact. Life preservers might also share some of the context of vests, while seat belts might share a context of belts, which isn't captured in the arithmetic above. But as long as there are no nearer terms, such differences can be treated as noise, for which the vectors are surprisingly robust. At some of these tasks the vectors perform very well: given a four-term analogy problem from the SAT with a multiple choice answer, such vectors can reach (Turney 2006) human performance. In other words, distributional semantic vectors automatically encode a great deal of common-sense about concepts in their structure.

Their ability to represent concepts as a whole makes them even able to handle some tasks that would normally be considered to require some creativity. For example, we performed the following experiment. All pairs of adjectives and nouns starting with the letter 'a', and pairs of rhyming words were generated based on existing word lists. The vectors representing both words in a pair were averaged, and then these were searched to find the nearest match to the vector for a search term. Here are synonymous alliterative phrases it came up with:

- robot: anthropomorphic automaton
- songbird: arboreal artist
- textbook: authoritative algebra

- birdhouse: architectural aviary
- chemistry: academic alchemy
- tin: antique aluminum
- bronze: archaeological alloy
- neon: amber ambiance
- divide: antagonistic arithmetic

and synonymous rhyming phrases:

- cowboy: colorado desperado
- friar: yeast priest, barbarian seminarian
- pillow: head bed
- trampoline: elastic gymnastic
- rocking chair: knitter sitter
- novel: fiction depiction
- orca: dalmation cetacean
- flower: bloom perfume, frilly lily
- Clinton: she nominee

The ability to generate such paraphrases demonstrates that some aspects of the meaning of the terms have been captured by the vectors. Other aspects of the meaning are missing, however. Learning solely from written texts creates some serious gaps in the knowledge implicitly contained in the set of vectors. Vectors nearby to the vector for **safe** include

- synonyms, near synonyms, hypernyms, sister terms, and hyponyms (**secure, healthy, reliable, comfortable, stable, protected, sane, adequate, prudent, reliable**)

- other forms of the word (**safer, safest, safely, unsafe, safeness, Safe, safety**)

- antonyms (**unsafe, dangerous, hazardous**)

This is not ideal. What we would like for the purposes of reasoning is a vector whose near terms are all synonyms of **safe**, with other related terms a little farther away**.** Perhaps most important is to draw a strong distinction in meaning between terms and their antonyms. One way to do this is to start with a vector space learned from a large text corpus, but then represent the concepts we are interested in as a sum or average of the synonomous terms that all mean the same concept. We could represent the concept of **safe**, for example, by a vector that is a weighted sum of the terms we would like nearby, with negative weights on the terms from which we would like it to be farther away. This separates the *concept* vectors from the vectors for the *terms* themselves, though the two will still be very close. The sets of synonyms and antonyms for many English terms are already available from Wordnet, so constructing such vectors is not difficult. (Rothe, 2015)

While a semantic vector space trained on a large, diverse text corpus is successful on some analogy tasks, it fails badly on others. For example, the following is an easy analogy for humans:

**blueberry : blue jay :: strawberry : cardinal**

If we try this in word2vec, instead of **cardinal** we get

**blue jays   red bellied woodpecker   grackle   ovenbird   downy woodpecker   indigo bunting   tufted titmouse   Carolina wren   chickadee   nuthatch   rose breasted grackle   bluejays   raccoon   spruce grouse   robin**

Most of the results are species of birds, but there does not seem to be any tendency for the birds named to be *red*. We can easily find clusters of fruit and songbirds in the vector space. But there is no reason for red things to appear in the same contexts in newspaper articles—no texts that discuss firetrucks, strawberries, and cardinals in the same way. Their color just is not very relevant to the way newspapers talk about those things. (If we trained the system on books for toddlers, that might be different.) Because red things are not clustered together, the analogy fails. The color of objects is not typically one of the facts about them that the vector encodes because it is too elementary a fact to be mentioned in the corpus.

In the last few years, computer vision researchers (Sadeghi 2015) (Reed 2015) (Upchurch 2016) have built deep-learning neural network architectures whose weights, treated as a high-dimensional vector, organize visual representations in the same way as distributional semantic vector spaces organize the textual world. These systems are able to form "visual analogies" that could potentially solve the analogy above, not in verbal form but with actual pictures of birds and fruit. Such systems are not trained on context at all (each image is learned in isolation) and yet they are able to build up a similar representation. This is because context serves mainly as a way of discovering similarity. If we can learn similarity directly, through the representations formed in a deep neural network, similar things (birds, fruits, items of particular colors or shapes) will be represented by similar vectors and form clusters, enabling the analogical arithmetic above to go through.

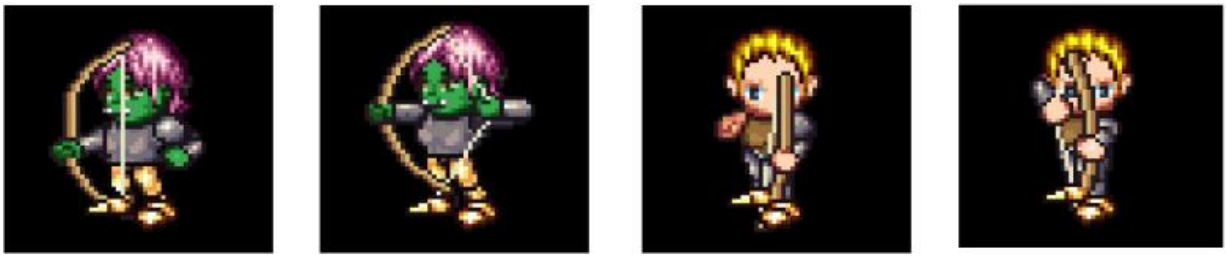

Figure: Example of deep visual analogy from (Reed 2015) Based on the first three images and a learned image manifold, their deep learning system is able to infer the fourth image by analogy with the other three.

*Any* method of creating high-dimensional vectors in order to classify data will have these properties of analogy-forming, relation-representation, and associational structure if it is successful in putting vectors nearby in the explicit and implicit categories we are interested in.

10.7 **What Needs to be Done**

The facts presented above give reasonable confidence that researching ways of building better semantic vector spaces and reasoning over them could prove successful. The area has seen a few high-profile advocates in the last few years. Hinton, now at Google, has been a proponent of encoding the meaning of entire sentences in a single vector in what are called "skip-thought" vectors (Ba 2016). Hawkins's (2007) memory-prediction framework also represents concepts as high-dimensional vectors, though his work concentrates on streams of information rather than single documents or images. Making such a system able to reason effectively, though, is still a challenging problem that will take many researchers some time to complete. Such a program would involve many different areas:

10.7.1  Learning more complex relations

The method for following chains of deductive reasoning described in section 10.5 is really only effective on one-to-one relations, where each concept in the input is paired with exactly one concept in the output. These include a wide variety of relations, everything from **scientist_studies_field (astronomer, astronomy)** to **animal_makes_noise(cow, moo)**. But most relations are not so simple. Consider the relation **store_carries_product**. Every store carries many products, and every product is carried by many stores. Associational reasoning is still helpful here— if a store carries peanut butter, there's a good chance that it also carries jelly—but it is not as simple as defining one vector defining the relation between Walmart and its set of products, subtracting out Walmart and adding Costco, and expecting that the set of products found will be very accurate. Using larger vectors, it is possible to define a vector representing a set of semantically similar things, especially if we also give information about where to make the cutoff between things in the set and things not in the set. However, some relations are simply going to be too complex to handle this way, and will need to be encoded by something more complex than a single vector, no matter how we arrange the rest of the concepts in the vector space. Such relations may need to be learned by a neural network, ideally in such a way that similar relations will be able to share information to develop into similar representations in the neural network.

There has been some exploration of how various relations are encoded into a distributional semantic vector space. Rei (2014) for example, explores how hyponyms swarm around a term. The possibility has also been explored of reshaping a vector space according to verified facts in (Faruqui 2014). One problem with doing this is that other relations, not explicitly included in such reshaping, may be distorted and so no longer have the analogical properties they had in the original vector space. Wang (2014) also explores putting a knowledge graph into a vector space.

10.7.2  Distributional semantics

Research into distributional semantic vector spaces has exploded in the last few years. Some interesting areas include choosing dependency based-word embeddings (Levy 2014), which modify the context window based on the sentence parse tree. GloVe and word2vec are among the most popular distributional semantic vector spaces at the moment due to their capacity for training on large corpora.

10.7.3  Semantics from images, video, and other data streams

Image vectors are derived from the weights of neural networks trained on images. Image vectors capable of capturing more than just class membership, but also encoding color, texture, pose, shape, and other information are still in their infancy. Some interesting efforts include Sadeghi (2015), Reed (2015), and Upchurch (2016). Similar vectors could potentially be derived from 3D sensors, whether depth-based or tactile, to provide another dimension of context.

### 10.7.4 Combining two vector spaces to better capture the knowledge learned from each

Simply concatenating the vectors from two spaces is one way to combine them, but there must be ways of knowing which kinds of facts have been captured better by one space than another, and giving more weight to that space. This seems like a task best handled by a neural network.

### 10.7.5 Encoding the meaning of natural language phrases and sentences as vectors

Skip-thought vectors (Kiros 2015) are one attempt at encoding sentences and phrases directly as vectors. AnalogySpace, built from the large knowledge base ConceptNet (Speer 2008), is another such example. If we keep to the idea of representing only word-like concepts as vectors so as to preserve the relational properties, a different approach would seem to be required. A sentence can be considered as a series of asserted relations between the terms in the sentence. If we can successfully parse the sentence into these relations, and modify our representations of the terms and relations appropriately, the system will have incorporated the knowledge in the sentence. Such semantic parsing is not completely reliable yet, but it has been improving. It would be best if such a semantic parsing system could be learned together with the semantic vector space so that more subtle conceptual relations could potentially be captured.

### 10.7.6 Modifying a semantic vector space as new information is learned without destroying already existing structure

(Faruqui 2014) showed that the vectors representing terms in a distributional semantic vector space could be modified to better capture certain known relations, but that doing so without regard to existing structure reduced the ability of the system to form analogies of other relations which were not explicitly optimized for. Unless these unplanned-for relations can be preserved somehow, modifying the vector space to incorporate known facts will always run the risk of destroying them.

### 10.7.7 Performing reasoning within vector spaces

When a chain of reasoning consists solely of one-to-one relations that are accurately captured by displacement vectors, deductive reasoning to follow the chain is a single-step process. This gives it a huge advantage over reasoning within a knowledge base, where a tree of possible relations must be explored to find a path between the terms in the relation to be proved. But this only applies to that limited set of one-to-one relations. For reasoning steps that involve more complex combinations of and/or operations on relations (known as Horn rules) or that involve higher-order relations, other techniques must be developed. This has not yet been extensively explored, although (Neelakantan 2015) and (Lin 2015) have made a beginning at it.

### 10.7.8 Ways of discovering and representing knowledge about physical consequences

Most of the concepts and relations that need to be represented in order to reason about consequences of actions will not come from either textual sources or from still images, but only through the experiences of an agent interacting with the world in a safe play environment. These experiences will need to incorporate spatial and temporal aspects which are difficult to handle in standard reasoning systems and vector-based ones alike. The kind of system outlined here at least has the capacity, though, to represent and reason about all the relevant concepts in a real-world situation which could involve all kinds of unanticipated objects and actors. The same can't be said of approaches using either a hand-built knowledge base or a learned neural-network representation that does not carry out reasoning processes.

## 10.8 Conclusion

Semantic vector spaces provide a way of capturing conceptual knowledge and reasoning about it efficiently and flexibly. They allow for analogical and associational reasoning in a way that is completely impractical for purely symbolic approaches to knowledge representation. Such conceptual representations are necessary for interacting with the diversity of situations that arise in the real world and doing so safely. There is still a lot of territory to explore in how best to create and make use of such subsymbolic representations, but research from neuroscience, machine learning, linguistic and knowledge representation communities all seem to be converging on similar methods, though what a finished system will look like is still murky. Any short-cuts to A.I. safety that are unable to reason about physical processes and human goals in order to reduce accidents and human error will behave in unexpected ways when introduced to new environments. But there is a road whose endpoint is a system that can understand what we want and take actions to achieve that, if we are willing to do the work necessary to follow it.